\documentclass[10pt, a4paper]{article}

\usepackage{lrec-coling2024} 
\usepackage{graphicx}
\usepackage{tabularx}
\usepackage{booktabs} 
\usepackage{siunitx} 
\usepackage{float}

\title{\textit{EcoVerse}: An Annotated Twitter Dataset for Eco-Relevance Classification, Environmental Impact Analysis,\\ 
and Stance Detection}

\name{Francesca Grasso, Stefano Locci, Giovanni Siragusa, and Luigi Di Caro} 

\address{Department of Computer Science, University of Turin\\
         Corso Svizzera 185 - 10149 Turin, Italy\\
         \{fr.grasso, stefano.locci, giovanni.siragusa, luigi.dicaro\}@unito.it\\}

\abstract{
Anthropogenic ecological crisis constitutes a significant challenge that all within the academy must urgently face, including the Natural Language Processing (NLP) community. While recent years have seen increasing work revolving around climate-centric discourse, crucial environmental and ecological topics outside of climate change remain largely unaddressed, despite their prominent importance. Mainstream NLP tasks, such as sentiment analysis, dominate the scene, but there remains an untouched space in the literature involving the analysis of environmental impacts of certain events and practices. To address this gap, this paper presents \textit{EcoVerse}, an annotated English Twitter dataset of 3,023 tweets spanning a wide spectrum of environmental topics. We propose a three-level annotation scheme designed for Eco-Relevance Classification, Stance Detection, and introducing an original approach for Environmental Impact Analysis. We detail the data collection, filtering, and labeling process that led to the creation of the dataset. Remarkable Inter-Annotator Agreement indicates that the annotation scheme produces consistent annotations of high quality.
Subsequent classification experiments using BERT-based models, including ClimateBERT, are presented. These yield encouraging results, while also indicating room for a model specifically tailored for environmental texts. The dataset is made freely available to stimulate further research.
 \\ \newline  }

\begin{document}

\maketitleabstract

\section{Introduction}
Climate change (CC) is both cause and consequence of environmental crisis and related critical events that have occurred in the last century and are gaining increasing impact on the
planet. Events that caused CC have been proven to be of anthropogenic origin \citep{Zwiers2000The,Storch2006Anthropogenic},
while the consequences affect entire ecosystems, humans, nonhumans, and nonliving
entities. As social sciences and humanities have long since already acknowledged, including the linguistics and ecolinguistics community \citep{stibbe2021ecolinguistics, Poole2022CorpusAssistedE
}, the call to action to address the ecological crisis is not solely directed at scholars directly linked to climate and environmental sciences. 
Ecolinguistic studies, for instance, have demonstrated that language use and narratives about the environment are crucial resources for understanding how individuals, as well as economic and media entities, perceive society and the natural world, and subsequently, how they act \citep{
Norton2019TellingOS,Zhang2023EcologicalDA}.

Within the Natural Language Processing (NLP) community, the past few years have witnessed an increase of interest towards CC and environmental topics.  
Most efforts in this domain have been primarily directed at CC discourse \citep{Stede2021TheCC}, with NLP techniques predominantly being applied for stance classification (e.g. \citet{Mohammad2016SemEval2016T6, Luo2020DeSMOGDS}), topic modelling (e.g. \citet{AlRawi2021TopicMO,Stede2023FramingCC}), and sentiment analysis (e.g. \citet{Dahal2019TopicMA, Mi2023TextMA}).
However, little attention has been given in the NLP field to investigating ecological-related issues from a broader perspective, where a more comprehensive exploration of the vast areas of environmental and ecological narratives are left largely untouched. 
Existing work 
that fall outside the scope of climate-centric discourse 
usually investigate 
public opinion or emotional responses towards very specific ecological events or phenomena (e.g. \citet{Duong2023NeurosymbolicAF,Roberts2018InvestigatingTE}). Among these works, very few datasets are available for computational analysis \citep{Ibrohim2023SentimentAF}. 
Moreover, there remains a notable gap in terms of tasks adopted, where the literature misses the analysis of textual content that reports or discusses events, actions, 
and beliefs having either beneficial or harmful impact for the environment and well-being of the natural world. Collecting such kind of data is of primary importance to examine the multifaceted relationships between human activities and ecological consequences, and to derive insights that can guide both policy-making and public awareness initiatives \citep{Drummond2018Is, Cecere2016Activities}.
To contribute to filling this gap, and as our primary contribution, in this paper we present the EcoVerse dataset\footnote{\href{https://github.com/GioSira/EcoVerse.git}{https://github.com/GioSira/EcoVerse.git}}, comprising 3,023 manually annotated English tweets and focusing on environmental and ecologically relevant topics. 
These range from biodiversity loss, textile waste, and
sustainable farming to plastic pollution, policy 
initiatives (e.g. COP26 discussions), and renewable energy. The tweets were annotated using a novel three-level annotation scheme, specifically designed to address three distinct objectives concurrently:\\
\textit{(i) Eco-Relevance classification}: A tweet is classified as either \texttt{eco-related} or \texttt{not eco-related}; \\\textit{(ii) Environmental Impact Analysis}: for eco-related tweets, this level determines whether the post conveys behaviors or events with beneficial, harmful/threatening, or neutral impacts on the environment. This level introduces a new paradigm of analysis, marking our second contribution;
\\\textit{(iii) Stance Detection}: The author’s stance is discerned and categorized as \texttt{supportive}, \texttt{neutral}, or \texttt{skeptical/opposing} towards environmental causes.

As a third contribution, we trained classification models on the annotated dataset and present the initial promising results.

To the best of our knowledge, EcoVerse represents the first available annotated dataset collecting textual instances across a wide spectrum of environmental topics, specifically developed for three different objectives.


The paper is structured as follows: Section \ref{relatedworks} provides an overview of the relevant related work. In Section \ref{datacollec} we detail the dataset creation process, while Section \ref{datasetannotation} describes the annotation scheme, procedure, results, and dataset statistics. Section \ref{experiments} presents and discusses
first classification results performed with BERT-based models, including ClimateBERT \citep{wkbl2022climatebert}. 
Finally, Section \ref{conclusion} concludes the paper and presents future directions.

\section{Related Works}
\label{relatedworks}

Climate and environmental awareness are gaining increasing interests in the AI community, including the NLP field. On one hand, climate change (CC) emerges as a framework for responsible research, seen in the rise of a new wave of research named \textit{green AI} or \textit{green NLP} \citep{Schwartz2019GreenA}. Such research is primarily aimed at addressing the environmental impact and computational costs of AI research \citep{Hershcovich2022TowardsCA} and encouraging reduction in resources spent, especially considering the enormous amount of energy used for training and running computational models \cite{Treviso2022EfficientMF}. On the other hand, CC has become the very subject of inquiry, with growing studies on climate discourse and debate.
\paragraph{NLP Methods and CC}

While there has been a surge in papers on CC texts, many converge on similar NLP tasks \citep{Stede2021TheCC}. We will spotlight the most relevant. Popular methods include stance detection, typically applied to distinguish "deniers" and "believers" \citep{Mohammad2016SemEval2016T6,Upadhyaya2022AMM,Vaid2022TowardsFC} or adopt more fine-grained distinction \citep{Anshelm2014DiscoursesOG, Luo2020DeSMOGDS}. Sentiment analysis is largely performed, typically in social media platforms such as Twitter \citep{MohamadSham2022ClimateCS}, usually to detect public opinions on the matter \citep{Dahal2019TopicMA,Koenecke2019LearningTU,Mi2023TextMA}.
Topic modelling (usually LDA \citep{blei2003latent})  is often applied to e.g. understand how social media users engage with discussions on CC (\citet{AlRawi2021TopicMO}; \citetlanguageresource{varini2020climatext}) or computing topic/sentiment correlations \citep{Dahal2019TopicMA,Jiang2017ComparingAT}.
\citet{Stede2023FramingCC} compute 'framing' categories that two editorials used, which were found to be relevant for CC.
\paragraph{Datasets on CC} Several efforts have been made to construct CC annotated datasets, although not all are publicly available. \citet{Effrosynidis2022TheCC} present one of the most extensive Twitter Dataset on CC discourse, 
with several dimensions of information tied to each tweet, including CC stance and sentiment, and topic modeling. The dataset is constructed by merging three datasets, two of which publicly available (
\citetlanguageresource{littman2019climate}; 
\citetlanguageresource{DVN/LNNPVD_2019}). \citet{Schaefer2022GerCCTAA} present a German CC Tweet Corpus 
for argument mining 
\citeplanguageresource{robin_schaefer_2022_6479492}. \citet{wkbl2022climatebert} perform three downstream 
tasks (text classification, sentiment analysis, and fact-checking) using ClimateBert, a language model for climate-related texts.
To do so, 
the authors employ a dataset of 
paragraphs from companies’ reports \citeplanguageresource{stammbach2023environmental} and 
ClimateFever \citeplanguageresource{Diggelmann2020CLIMATEFEVERAD}, containing 
sentences that make a claim about climate-related topics. \citet{Bartsch2023TheIC} constructed a multimodal dataset on CC discourse.

\paragraph{Environmental Dicourse}

In terms of ecological 
narratives that extend beyond the CC debate \textit{stricto senso}, there exists a limited body of work, and even fewer datasets.
A 
survey from \citet{Ibrohim2023SentimentAF} on sentiment analysis for environmental topics shows that only one of the annotated datasets for model training and testing is open to researchers 
\citep{Hagerer2021EndtoEndAB}. Moreover, existing studies tend to embark on isolated investigations \citep{Duong2023NeurosymbolicAF}, seldom painting a comprehensive picture of broader environmental narratives. As a result, numerous "green" topics remain completely unexplored.
Or, they might narrow their focus to a specific target audience \citep{Serrano2020ExploringPA}, and usually undertake mainstream NLP tasks. Among few available datasets, \citetlanguageresource{manifestoproj2023} 
analyses parties’ manifestos, including three ecology-related categories in their annotation scheme. 
In this scenario, a comprehensive exploration of diverse environmental topics is lacking, as is any analysis of the ecological impact of specific events.
Our proposed dataset, EcoVerse, aims to fill this gap 
by proposing a publicly available Twitter 
collection for Eco-Relevance, Environmental Impact Analysis, and Stance Detection. 

\section{Data Collection and Cleaning}
\label{datacollec}

\subsection{Data Collection}

We chose Twitter\footnote{
During the realization of this project, Twitter rebranded to 'X'.  Due to the change's recency and the original name's widespread recognition, we continue referring to it as 'Twitter' in this article.} as our primary source for data extraction, primarily due to its convenience and alignment with our objectives. The platform's high popularity ensures access to a diverse user base, ranging from individual users to sectorial magazines and media outlets. Given the recent surge in discussions on climate change (CC) and environmental issues, Twitter presents a rich source of such conversations. Moreover, Twitter's advanced API\footnote{\url{https://developer.twitter.com/en/products/twitter-api}} was conducive to our precise data collection needs. 
The data extraction criteria were carefully elaborated to align with our annotation objectives, ensuring balance and heterogeneity within the dataset. Our goal was to curate tweets potentially classifiable under varied categories: eco-related or not, with content having negative, positive, or neutral environmental impacts, and representing diverse stances - supportive, neutral, or opposing - towards environmental actions.
To this aim, we employed Twitter hashtags to crawl tweets that might hold eco-related content with a spectrum of stances and environmental impacts. In particular, the tag \texttt{\#environment} was used for general eco-related content, whereas \texttt{\#climatescam} and \texttt{\#ecoterrorism} targeted tweets from potentially skeptical users.
To ensure inclusion of non eco-related content, we crawled tweets from mainstream news sources (i.e. \texttt{@telegraph}, \texttt{@nytimes}, \texttt{@business}). We also mined tweets from popular environmental 
organizations and publications (i.e. \texttt{@natgeo}, \texttt{@NatGeoMag}, \texttt{@NatGeoPR}, \texttt{@Sierra\_Magazine}, \texttt{@nature\_org}) to ensure a wide array of content related to environmental issues. This choice also guaranteed the presence of nuanced polarities in terms of ecological content, intentionally including material that could be deemed "borderline" (i.e. straddling categories). Such content allows for a more refined analysis in our annotation due to its inherent ambiguity. 
The data extraction spanned tweets from January 2019 to June 2023. The beginning of 2019 was strategically chosen, given its significance with the emergence of global movements like Fridays For Future, which gave a considerable boost to discussions on CC and environmental topics on the platform.

\subsection{Data Cleaning and Dataset Creation}
\label{datacleaning}
After collecting the tweets, we carried out several pre-processing steps. We started with basic filtering where we removed retweets, duplicates, and non-English tweets. Additionally, tweets containing only emojis, hashtags, tags, or URLs were excluded. Considering content length, we removed posts with fewer than 24 words, as they might contain insufficient information for meaningful analysis \citep{Yang2017UsingWE}. For content formatting, line breaks were eliminated, and links within the tweets were replaced with a generic "[URL]" placeholder. 
Following \citet{Lee2021DeduplicatingTD}, we utilized the Python library \texttt{MinHash LSH}\footnote{\url{http://ekzhu.com/datasketch/lsh.html}} 
to address redundancy 
of tweets.
By configuring the \texttt{threshold} parameter to a value of 0.2, we systematically eliminated tweets exhibiting a similarity score equal to or exceeding 20\%.
Post-cleaning, the dataset contained 21,244 unique tweets eligible for manual annotation. To streamline our data processing, we grouped the tweets into four "buckets" based on their sources, whith each bucket corresponding to the general categories mentioned in the previous section: 
"Environmental organizations and publications"; "Likely not eco-related"; "Likely eco-related"; "Likely skeptical". Grouping the tweets in this manner was instrumental in facilitating a balanced dataset during its creation. 
Out of the 21,244 tweets, we selected a final sample of 3,000 tweets. This size was determined based on the feasibility of manual annotation while ensuring the dataset's representativeness for our classification tasks. 
The 3,000 tweets were selected to ensure balance within the dataset, aiming for an equal distribution of eco-related and non eco-related tweets and a diverse representation across the other two annotation levels.
Accordingly, our tweet selection followed specific ratios: 33.34\% from the ”Environmental organization and publication” sources bucket; 26.67\% from the ”Likely not eco-related” sources bucket; 26.66\% from the ”Likely eco-related” bucket; and 13.33\% from the ”Likely skeptical” category. For each bucket, tweets were selected uniformly and randomly from each contributing source, ensuring an unbiased representation across all sources within that bucket, while also attempting to capture a diverse diachronic distribution over the time span of the tweets.


\section{Dataset Annotation}
\label{datasetannotation}

\subsection{Annotation Scheme and Guidelines}
\label{annotationscheme}

For the annotation of our 3k English tweets 
we developed a unique three-layer annotation scheme. Each layer corresponds to a distinct level of analysis. Below, we provide a concise overview of each level, mirroring the guidelines provided to the annotators. The first and second level of the annotation scheme drew inspiration from the schemes described in \citet{wkbl2022climatebert}, which were originally set up for text classification and sentiment analysis. Adapting their "climate-related" versus "not climate-related" taxonomy, we widened the scope to a more extensive environmental context. In terms of sentiment, while the authors identify climate-related sentiments as negative risks, positive opportunities, or neutral, we developed a new framework for Environmental Impact Analysis. For the third level, 
while inspired by conventional CC-stance detection tasks (as discussed in Section \ref{relatedworks}), we tailored our approach to encompass the broader range of environmental causes and concerns within the tweets.


    

\begin{description}
    \item[Eco-Relevance] This level aims to differentiate between posts that are relevant to ecology and environmental issues and those that are not, or those that present such topics descriptively (i.e. if a post provides a simple description of animals or plants or does not pertain to ecology, it would be categorized as not ecology-related). The labels for this binary classification are: \texttt{eco-related} / \texttt{not eco-related}. Only if a tweet is eco-related, do the subsequent annotation levels apply.


    \item[Environmental Impact] This level is designed to determine whether a tweet reports on behavior, practices, events, beliefs, or attitudes that might have a beneficial, harmful/threatening, or neutral impact on the environment.
    For instance, if a tweet discusses topics that are either explicitly or potentially harmful to ecological sustainability or the well-being of the natural world, it would be labeled as \texttt{negative}. Conversely, tweets portraying topics beneficial to the environment would be labeled \texttt{positive}. Posts that do not clearly fall into either category are marked as \texttt{neutral}. Examples can be found in Table \ref{tab:tweet_examples}.
    Importantly, after the initial pilot annotation (
    described in the subsequent section), we clarified in the guidelines that 
    even for eco-related tweets, the second-level annotations can be conditional and remain untagged if it is not possible to annotate them. 
    This scenario arises when such tweets do not distinctly represent or convey actual, clear, or directly identifiable events or practices.
    This is the case, for instance, of tweets voicing general opinions about environmental causes or presenting misinformation (e.g. \textit{"The greenies' only solution seems to be halting progress, reducing the population, and imposing communism with their climate cult"}). Nonetheless, the proportion of tweets left untagged at this level is minor and has not compromised the richness of the dataset or the results of classification experiments, as showed in the following sections.    
    \item[Stance Detection] The final level discerns the author's stance towards the environmental topic at hand. 
    A \texttt{supportive} stance shows concern towards the environmental crisis and related critical events and/or endorsement of ecological causes or practices that combat CC, environmental protection, or other sustainable behaviors. A \texttt{neutral} stance presents information without taking a clear position or expressing a specific viewpoint towards the ecological cause. Meanwhile, a \texttt{skeptical/opposing} expresses skepticism or denial towards the ecological cause, questioning the severity of CC or dismissing the need for environmental measures. They may also report disruptive behavior as good or normalize such behavior.
    
\end{description}

In Table \ref{tab:tweet_examples} we report some examples of tweets with their expected label, according to our developed 
annotation scheme.

\begin{table*}[ht]
\centering
\footnotesize
\begin{tabular}{|p{8.2cm}|p{2.9cm}|p{1.5cm}|p{1.8cm}|}
\toprule
\textbf{Tweet examples} & \textbf{Eco-Relevance} & \textbf{Environm. Impact} & \textbf{Stance} \\
\toprule
Just read an article on deforestation in the Amazon. The rate at which we are losing our forests is alarming. We need stricter regulations. & \texttt{eco-related} & \texttt{negative} & \texttt{supportive} \\
\midrule
Nefertiti's allure isn't just about her beauty. Recent Egyptology studies have started delving deeper, exploring her significant roles as co-pharaoh and regent in history. & \texttt{not eco-related} &  &  \\
\midrule
They say organic farming has reduced soil erosion and curbed water pollution. But to me? It's just another overpriced trend capitalizing on green hysteria. & \texttt{eco-related} &  \texttt{positive} & \texttt{skeptical/ opposing} \\
\midrule
The World Conservation Union's latest report states that over 1,000 mammal species are now classified as endangered due to habitat loss and poaching. & \texttt{eco-related} & \texttt{negative} & \texttt{neutral} \\
\midrule
South Africa leads in low residue levels in food, thanks to its climate and sustainable practices. A top choice for eco-conscious supply. \#environment \#fruitindustry  & \texttt{eco-related} & \texttt{positive} & \texttt{supportive} \\ \midrule
Three species of elephants are able to live in very different environments on
two continents, thanks to this stunning set of adaptations. & \texttt{not eco-related} &  &  \\
\bottomrule
\end{tabular}
\caption{Examples of tweets with expected labels. }
\label{tab:tweet_examples}
\end{table*}

 \begin{figure*}[!ht]
\begin{center}
\includegraphics[scale=1.0]{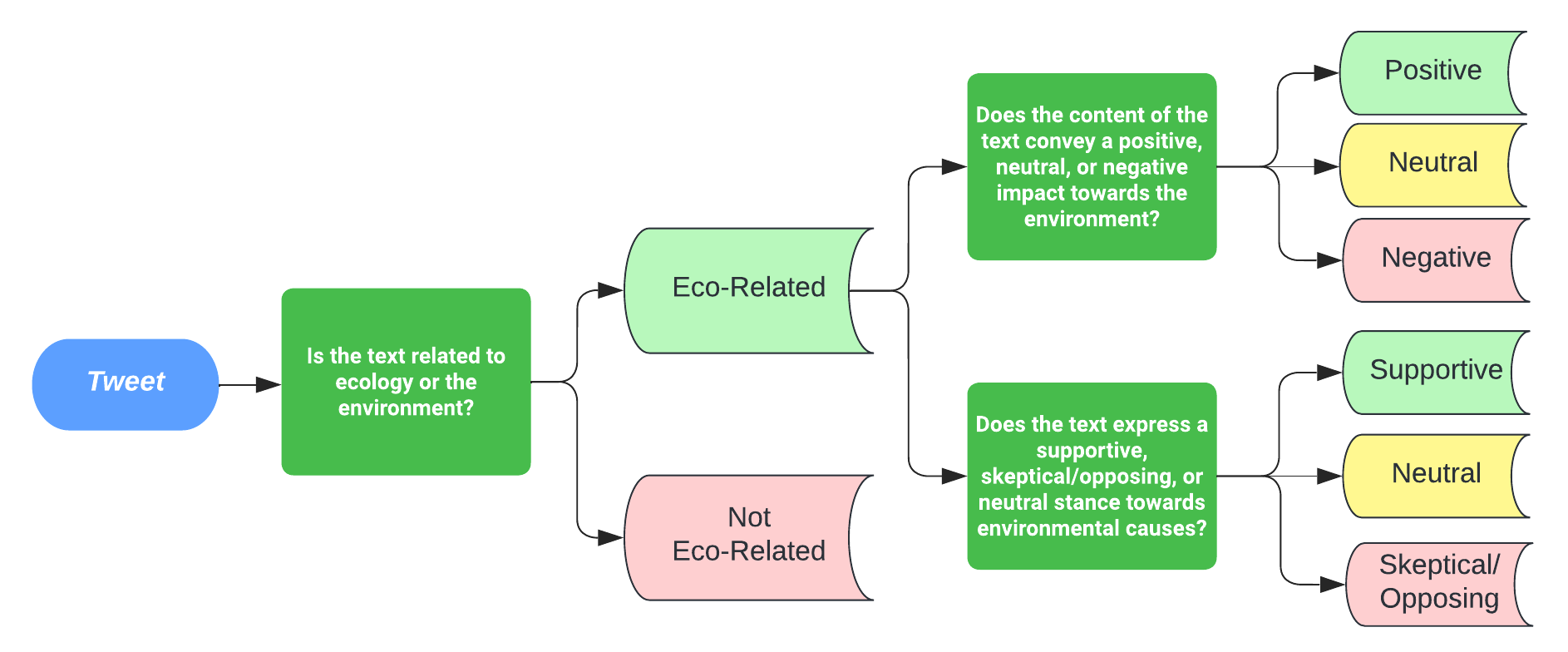} 
\caption{The annotation begins with tweet text analysis, assessing its ecological relevance. If the tweet is categorized as \texttt{eco-related}, subsequent steps involve Environmental Impact Analysis and Stance Detection. Conversely, for tweets deemed \texttt{not eco-related}, annotations are finalized and move to the next tweet.
}
\label{Fig: flowchart}
\end{center}
\end{figure*}

\subsection{Annotation Procedure}
The dataset annotation process was conducted by two annotators, who are also co-authors of this paper. Their expertise in environmental domains ensured a deep understanding of the data content and its nuances. While they have backgrounds in either computer science or linguistics, their primary strength lies in their robust understanding of ecological and environmental topics.
Accordingly, they were trained to perform the annotation task, which consists of three subtasks for each tweet:
1) Determine Eco-Relevance; for the eco-related tweets, 2) Determine Environmental Impact; 3) Identify the Stance. 
The annotation was carried out using the Label Studio open-source data labeling tool, which provides a web-enabled dynamic graphical interface
\footnote{\url{https://labelstud.io}}. Figure \ref{Fig: flowchart} illustrates the annotation process flowchart.
The annotators were provided with detailed guidelines that started with an introduction, offering both background and motivation. Additionally, when analyzing the tweets, annotators were encouraged to make inferences based on their background knowledge of "green" and environmental topics. The guidelines also clarified that certain annotations could be skipped (via a specific "skip" button provided by the labeling tool interface) if a tweet did not meet the criteria required for annotation (e.g., not written in English, contains only emojis, etc.) or if there was uncertainty about how to annotate a critical tweet. These would be revisited and discussed with the other annotator at the end of the task.
To ensure a shared understanding of the guidelines and consistency between the annotators, an iterative two-step training was undertaken. Initially, a pilot annotation was performed on a secondary dataset of 50 tweets (
with the tweet distribution as in the primary 3k dataset, described in Section \ref{datacleaning}). They compared their results, discussed different interpretations in order to gain familiarity with the scheme, solve open questions, and eventually improve the annotation guidelines. This process was repeated until they both felt comfortable in completing the main task.
Subsequently, they annotated an initial batch of 2,000 tweets from the primary dataset. After this, we monitored the Inter-Annotator Agreement (IAA), and annotators discussed any problematic tweets 
and possible refinements to the guidelines. With satisfactory IAA scores and most issues addressed,  
the annotation of the remaining tweets proceeded without further discussion sessions. 
After completing the main task, the annotators addressed the "skipped" tweets. They agreed on which ones to exclude, leading to an additional batch of 230 tweets being annotated to reach the desired dataset size of 3k tweets (once again, this batch's tweet selection followed the primary dataset's distribution ratio). 
Once the additional tweets were annotated, the annotators repeated the discussion and alignments on the "skipped" tweets. After that, the total count for the final dataset reached 3,023 annotated tweets.
\subsection{Annotation Results}
We calculated Cohen’s Kappa scores to measure Inter-Annotator Agreement (IAA) \citep{Artstein2008InterCoderAF}. Initially, IAA was calculated post the primary annotation task on the final annotators' datasets. Following this, the annotators engaged in a last in-depth discussion addressing disagreements across the three annotation levels. This exercise aimed not only to resolve these disagreements but also to rectify any potential oversight errors, ensuring the reliability of the annotations. Post this discussion, a subsequent IAA measurement was taken. Table \ref{tab:iaa_discussion} illustrates the IAA scores before and after this discussion.

\paragraph{Initial IAA measurement.} The Eco-Relevance Classification task yielded an initial IAA score of an impressive 0.85. This is consistent with the binary nature of the task and its relatively straightforward classification process. Moreover, this high score might underscore the clarity of the task guidelines.
With a more nuanced granularity, the annotation of the Stance showed an IAA of 0.79, slightly lower than the Eco-Relevance task. The slight drop is understandable, given that certain tweets potentially straddle categories, especially between supportive/neutral or skeptical-opposing/neutral.
Environmental Impact Analysis had an IAA of 0.67. This lower score can stem from both the dichotomy between tagged and untagged instances, and the inherent subjectivity of certain tweets. 
Notably, some tweets presented dual perspectives, such as highlighting protective environmental measures while simultaneously indicating an ongoing detrimental condition.
\paragraph{Addressing Disagreement.} To streamline the reconciliation process, the annotators first tackled disagreements in the Eco-Relevance Classification, as conflicts here would inevitably cascade to the subsequent levels. For example, a \texttt{not eco-related} label meant no annotations for the following levels. They then addressed discrepancies in Environmental Impact Analysis, starting with instances where one annotator's tagging was absent. Discussions prioritized "strong" mismatches, i.e., cases where one annotator marked a tweet as \texttt{positive} while the other labeled it \texttt{negative}. Disagreements between \texttt{neutral} and the other two labels weren't as intensively debated. The Stance Detection level was the last to be addressed, focusing on stark disagreements, particularly between \texttt{supportive} and \texttt{skeptical/opposing} labels.

Throughout these discussions, tweets were deeply analyzed to reach consensus wherever feasible. Taking a cue from \citet{Basile2021WeNT}, we also acknowledged that certain disagreements were “acceptable” due to the inherent subjectivity of some tweet content.
Some disagreements arose from simple annotators' oversight, but these were the minority.
Post-discussion, the IAA scores witnessed significant improvements, with 0.94 for Eco-Relevance classification, 0.82 for Environmental Impact Analysis, and 0.86 for Stance Detection. This improvement suggests that, after resolving any misunderstandings or discrepancies, the annotators were largely in agreement, which bodes well for the dataset's reliability.




\begin{table}[h]
    \centering
    \begin{tabular}{lcc}
        \toprule
        Task & \multicolumn{2}{c}{Cohen's $\kappa$} \\
        & Pre-Disc. & Post-Disc. \\
        \midrule
        Eco-Rel. Classification & 0.8507 & 0.9371 \\
        Env. Impact Analysis & 0.6705 & 0.8182 \\
        Stance Detection & 0.7868 & 0.8599 \\
        \bottomrule
    \end{tabular}
    \caption{Inter-Annotator Agreement (IAA) for each annotation task, before and after discussing disagreements.}
    \label{tab:iaa_discussion}
\end{table}

\subsection{Dataset Statistics}

\begin{table*}[th!]
    \centering
    \begin{tabular}{lcccccccc}
        \toprule
        & \multicolumn{2}{c}{Eco-Relevance} & \multicolumn{3}{c}{Env. Impact Analysis} & \multicolumn{3}{c}{Stance Detection} \\
        \cmidrule(lr){2-3} \cmidrule(lr){4-6} \cmidrule(lr){7-9}
        Annotator & Eco-rel. & Not eco-rel. & Pos. & Neu. & Neg. & Supp. & Neu. & Skep./Opp. \\
        \midrule
        Annotator I & 1,559 & 1,464 & 501 & 369 & 416 & 887 & 305 & 367 \\
        Annotator II & 1,530 & 1,493 & 629 & 221 & 457 & 887 & 327 & 316 \\
        \bottomrule
    \end{tabular}
    \caption{Labels distribution through the dataset for both Annotator I and Annotator II.}
    \label{tab:dataset_stats}
\end{table*}

Table \ref{tab:dataset_stats} reports the label distributions across the three annotation levels for both Annotator I and II. For the Eco-Relevance level, labels are well balanced, with a near even distribution between \texttt{eco-related} and \texttt{not eco-related}. On the Environmental Impact level, there is a balanced distribution, though Annotator II shows a slight inclination towards the \texttt{positive} label. 
For the Stance level, there is a noticeable majority of \texttt{supportive} labels. This is largely due to the nature of tweets reporting environmental news, events, or behaviors. Whether they reference beneficial or harmful environmental instances, they are typically conveyed in a manner that emphasizes concern, awareness, or advocacy for ecological well-being, hence leading to a supportive stance.
Table \ref{tab:text_stats} offers a succinct overview of essential statistical characteristics of the textual content of the 3,023 tweets. Notably, the mean count of hashtags per tweet, approximately 2, with a standard deviation of $\pm$ 3, indicates a relatively restrained usage of hashtags within the dataset, with a modest degree of variability. Moreover, after the removal of stop words, the dataset exhibits a total of 56,115 tokens presenting 11,026 unique words (Types).
Finally, the mean length and the standard deviation of tweets is provided for a comprehensive understanding of the distribution and variation in text lengths within the dataset.

\begin{table}[h!t]
\centering
\begin{tabular}{lc}
\toprule
\textbf{Statistic} & \textbf{Value} \\
\midrule
Num. of hashtags & 5,442 \\
Avg. hashtags per tweet & $\sim$ 2 $\pm$ 3 \\
Avg. tweet length (words) & $\sim$ 43 $\pm$ 10 \\
Num. of tokens w/o stopwords & 56,115 \\
Num. of types |$\nu$| w/o stopwords & 11,026 \\
\bottomrule
\end{tabular}
\caption{Summary of Dataset Statistics}
\label{tab:text_stats}
\end{table}

\paragraph{Environmental Topics}
To provide a comprehensive overview of the range of environmental topics captured in EcoVerse and emphasize its diversity, we performed a manual extraction of frequently observed keywords and clustered them into distinct topics\footnote{To determine both keywords and topics, we first took inspiration from Worcester Polytechnic Institute's topic list (\url{https://libguides.wpi.edu/sustainability}) and \citet{Ibrohim2023SentimentAF}, then expanded on them.
}. Twitter data is characterized by unique elements such as hashtags, mentions, and abbreviations. Additionally, tweets often encompass context-dependent nuances like irony, humor, and sarcasm. These characteristics make it particularly challenging for automated algorithms, which typically rely on term frequency and document co-occurrence patterns. Given the context-rich nature of tweets, the risk of misinterpretation by automated methods is heightened \citep{infExtSM}. On the other hand, manual extraction leverages human expertise and contextual understanding, allowing for a deeper understanding of the nuances often overlooked by automated techniques.
Table \ref{tab:topics} report the environmental topics identified in EcoVerse.

\begin{table}[h!bt]
    \centering
    \begin{tabularx}{\columnwidth}{Xr}
    \toprule
        \textbf{Topic} & \textbf{\%} \\
    \midrule    
        Fashion (un)Sustainability & 0.3\% \\
        Melting Ice and Sea Level Rise & 0.5\% \\
        Agriculture & 0.8\% \\
        Environmental and Climate (in)Justice & 0.9\% \\
        Food (un)Sustainability and Safety & 1.5\% \\
        Plastic and other Pollutants & 1.6\% \\
        Water Management and Conservation & 2.6\% \\
        Environmental Activism & 2.8\% \\
        Others & 3.7\% \\
        Eco-Practices and Circular Economy & 4.2\% \\
        Air Pollution and Emissions & 5.8\% \\
        Environmental Initiatives & 6\% \\
        Environmental Policy and Agreements & 7\% \\
        Deforestation and Land Management & 7.1\% \\
        Energy Sources and Consumption & 8.3\% \\
        Sustainability (in general) & 8.8\% \\
        Biodiversity and Ecosystem & 10.1\% \\
        Climate Change & 11\% \\
        Ecology and Environment (in general) & 17\% \\
   \bottomrule
   \end{tabularx}
    \caption{Environmental topics identified in EcoVerse.}
    \label{tab:topics}
\end{table}


\section{
Experimental Setup}
\label{experiments}

We performed a comprehensive evaluation to automatically identify the three levels within the annotation scheme: Eco-Relevance, Environmental Impact, and Stance Detection. This assessment serves two main purposes: to rigorously gauge the proficiency of state-of-the-art models in recognizing these levels and to establish benchmarks for future models.
\
For our experiments, we used the dataset annotated by Annotator II, cleaned of ”[URL]” placeholders and user mentions (defined by the prefix @). The dataset was split into training, evaluation, and test sets with a 80-10-10 ratio.

        
        

We fine-tuned six pre-trained BERT-based models \citep{devlin-etal-2019-bert,liu2020roberta,sanh2019distilbert} from the Hugging Face platform\footnote{\url{https://huggingface.co/}}.
All models were trained on an NVidia T4 GPU and took about 5 minutes each to converge. We set the learning rate to $3\times10^{-5}$, the batch size to 16, and the number of maximum epochs to 10. We used AdamW \cite{loshchilov2018decoupled} optimizer with an epsilon of  $2\times10^{-10}$.  All the hyperparameters were found via a grid search on the evaluation set.
The models include:\\
\textbf{BERT and RoBERTa}: Masked language models trained on large English corpora. 
We fine-tuned \texttt{bert-base} and \texttt{roberta-base}, which act as lower bounds for our tests.
\\\textbf{DistilRoBERTa}: 
A faster, more compact version of RoBERTa, pre-trained on the same corpus. Since \citet{wkbl2022climatebert}'s ClimateBERT models are based on DistilRoBERTa, we fine-tuned the \texttt{distilroberta-base} to assess potential enhancements for ClimateBERT.
\\\textbf{ClimateBERT}: 
We fine-tuned 
\texttt{$ClimateBert_F$}, \texttt{$ClimateBert_S$} and \texttt{$ClimateBert_{S+D}$}. The models are trained on subsets of \citetlanguageresource{stammbach2023environmental}'s dataset. 


\subsection{Evaluation}
\label{evaluation}

To assess the performance of these models, we employed metrics including Accuracy, Precision, Recall, and F-measure. 
Using Scikit-learn's implementation, we set the average to both \texttt{micro}
 (\texttt{m}) and \texttt{macro} (\texttt{M}). Notably, in multi-class settings, micro-average Precision, Recall and $F1$ scores are all identical to Accuracy. Hence, only Accuracy micro measure $A_m$ is shown in the results.
Table \ref{tab:ecorel} summarizes our findings for the Eco-Relevance task. DistilRoBERTa stands out in accuracy, effectively identifying eco-relevant tweets, while $ClimateBert_F$ lags behind RoBERTa and DistilRoBERTa. The specialized climate knowledge in $ClimateBert_F$ does not offer a significant advantage for this task. BERT shows the lowest performance among the classifiers. For the macro F-measure, RoBERTa leads, followed by $ClimateBert_{S+D}$, with $ClimateBert_F$ being the least effective.


\begin{table}[h!t]
    \centering
    \begin{tabular}{
        l
        S[table-format=2.2]
        S[table-format=2.2]
        S[table-format=2.2]
        S[table-format=2.2]
    }
    \toprule
    {Model} & {$A_m$} & {$P_M$} & {$R_M$} & {$F1_M$} \\
    \midrule 
        $BERT$ & 85.8 & 87.51 & 87.88 & 87.65\\
        $RoBERTa$ & 88.87 & \bfseries 88.90 & \bfseries 88.96 & \bfseries 88.93\\
        $D.RoBERTa$ & \bfseries 89.43 & 88.33 & 88.12 & 88.22\\
        $ClimBert_F$ & 86.47 & 86.57 & 86.57 & 86.57\\
        $ClimBert_S$ & 88.12 & 87.91 & 87.91 & 87.91\\
        $ClimBert_{S+D}$ & 87.79 & 88.23 & 88.30 & 88.26\\
    \bottomrule
    \end{tabular}
    \caption{Results on the Eco-Relevance task. $A_m$ stands for Accuracy, $P_M$ for Precision, $R_M$ for Recall, and $F1_M$ for F-measure. $ClimBert$ and $D.RoBERTa$ are the ClimateBert and DistilRoBERTa models, respectively.}
    \label{tab:ecorel}
\end{table}

In Table \ref{tab:sentiment}, we present the results for the Environmental Impact task. Analyzing Accuracy, $ClimateBert_S$ leads with 78.62\%, closely followed by $ClimateBert_{S+D}$ at 77.24\%. RoBERTa is not far behind, with an Accuracy of 76.55\%.
The metrics $P_M$, $R_M$, and $F1_M$ provide a comprehensive view of the models' performance. BERT tops in Precision with 62.08\%. However, it is important to note the significantly low $F1_M$ scores of both BERT and $ClimateBert_{S+D}$, hovering around or below 30\%. Upon employing a confusion matrix for a deeper analysis, it was observed that the models encountered difficulties in accurately distinguishing the \texttt{neutral} label. Reflecting the Accuracy analysis, $ClimateBert_S$ stands out, followed by DistilRoBERTa with an F-measure of 54.67\%.

\begin{table}[h!tb]
    \centering
    \begin{tabular}{
        l
        S[table-format=2.2]
        S[table-format=2.2]
        S[table-format=2.2]
        S[table-format=2.2]
    }
    \toprule
    {Model} & {$A_m$} & {$P_M$} & {$R_M$} & {$F1_M$} \\
    \midrule
    $BERT$              & 73.79 & \bfseries 62.08 & 35.37 & 30.37 \\
    $RoBERTa$           & 76.55 & 53.17 & 54.11 & 52.68 \\
    $D.RoBERTa$     & 74.24 & 50.73 & \bfseries 57.90 & 53.87 \\
    $ClimBert_F$      & 76.55 & 48.85 & 55.37 & 51.82 \\
    $ClimBert_S$      & \bfseries 78.62 & 51.81 & 57.78 & \bfseries 54.67 \\
    $ClimBert_{S+D}$  & 77.24 & 52.81 & 35.46 & 28.59 \\
    \bottomrule
    \end{tabular}
    \caption{Results on the Environmental Impact Analysis task.}
    \label{tab:sentiment}
\end{table}


For the Stance Detection task, Table \ref{tab:stance} summarizes the results. Both RoBERTa and DistilRoBERTa excel in detecting stances related to environmental topics, as evidenced by their elevated $A_m$ scores. This underscores their effectiveness in handling nuances of environmental discourse. Conversely, the ClimateBert models show a slight decline in performance, with drops of 11.69\% for $ClimateBert_F$, 8.78\% for $ClimateBert_S$, and 5.85\% for $ClimateBert_{S+D}$ compared to BERT and RoBERTa.
In terms of the macro F-measure, BERT outperformed all models, with RoBERTa following closely. However, $ClimateBert_{F}$ and $ClimateBert_{S+D}$ show the lowest scores.

\begin{table}[h!t]
    \centering
    \begin{tabular}{
        l
        S[table-format=2.2]
        S[table-format=2.2]
        S[table-format=2.2]
        S[table-format=2.2]
    }
    \toprule
    {Model} & {$A_m$} & {$P_M$} & {$R_M$} & {$F1_M$} \\
    \midrule
        $BERT$ & 74.27 & \bfseries 95.09 & \bfseries 96.04 & \bfseries 95.56\\

        $RoBERTa$ & \bfseries 81.29 & 94.61 & 94.61 & 94.61\\
        $D.RoBERTa$ & \bfseries 81.29 & 92.81 & 92.81 & 92.81\\
        $ClimBert_F$ & 69.6 & 92.29 & 91.38 & 91.83\\
        $ClimBert_S$ & 72.51 & 94.12 & 93.18 & 93.64\\
        $ClimBert_{S+D}$ & 75.44 & 91.57 & 92.44 & 92.00\\
    \bottomrule
    \end{tabular}
    \caption{Results on the Stance Detection task. 
    }
    \label{tab:stance}
\end{table}

\subsection{
Addressing \texttt{\#climatescam} Bias
}
During the annotation process, we observed that most tweets labeled as \texttt{skeptical/opposing} in the Stance Detection task included the hashtag \texttt{\#climatescam}. To prevent a potential bias on such hashtag, where the neural network might over-rely on this one for classification, we decided to conduct a comparative analysis using a dataset where this hashtag has been removed. 
Table \ref{tab:noclimscam_general} presents outcomes for the top-performing models for each classification task on the dataset without the \texttt{\#climatescam} hashtag. Notably, the Eco-Relevance and Environmental Impact tasks see accuracy improvements of 0.33\% and 0.69\%, respectively. As expected, the most significant variance appears in the Stance Detection task. Excluding \texttt{\#climatescam} led to a 4.69\% drop in performance, highlighting the hashtag's pivotal role in identifying Skeptical/Opposing tweets.
We also experimented by removing \texttt{\#environment}, another frequent hashtag present in the dataset, obtaining negligible results. 

\begin{table}[h!t]
    \centering
    \begin{tabular}{l p{2.2cm} c}
    \toprule 
    Task & Model & $A_m$ \\
    \midrule 
    Eco-Relevance & $D.RoBERTa$ & 89.76\\
    Environmental Impact & $ClimateBert_S$ & 79.31 \\
    Stance Detection & $D.RoBERTa$ & 76.60 \\
    \bottomrule 
    \end{tabular}
    \caption{Results of the most performant models trained without \texttt{\#climatescam} for each task.}
    \label{tab:noclimscam_general}
\end{table}

\section{Conclusion and Future Work}
\label{conclusion}
In this paper, we presented EcoVerse, an English Twitter dataset with 3,023 annotated tweets encompassing a wide spectrum of environmental topics. 
We propose a novel three-level annotation scheme, designed for Eco-Relevance Classification, Environmental Impact Analysis, and Stance Detection. The second constitutes an original paradigm to evaluate the relationships between events and human activities and their ecological consequences.
High IAA scores and 
well-balanced labels distribution 
demonstrate the dataset's reliability. 
Experiments with BERT-based models, including ClimateBert, 
indicate the potential for specialized language models for environment-related texts, which constitute the first of our future work. 
We also plan to expand the dataset, 
and try classification tasks on non-annotated data. 
EcoVerse aims to serve as a valuable tool for both researchers and stakeholders in the environmental field, whether for policy insights, awareness initiatives, or to encourage environmental discourse exploration and research.

\section{Acknowledgements}
We would like to express our gratitude to Almawave S.p.A. for their substantial technical and logistical support throughout the development of this project.

\section{Ethical Statement}
\label{Ethical statement}
In this study, we acknowledge the significant ethical considerations surrounding the use of deep neural networks. As stewards of the environment, we recognize that the computational power required for training these models can result in substantial carbon emissions, thereby contributing to climate change. To address this concern, we have undertaken a conscientious approach by quantifying the carbon footprint associated with our training process. To achieve this objective we employed the Python library \texttt{CodeCarbon}\footnote{\url{https://pypi.org/project/codecarbon/}}, which provides a reliable framework for measuring the carbon footprint of computing workloads. We recorded and analyzed the emissions data throughout the training and hyperparameter optimization phases. In Table \ref{table:carbon-emissions}, we present the details of the estimated Co2 emissions released during the entire training process. In total, the training of six models on the EcoVerse dataset, encompassing the process of hyperparameter tuning and including all the experiments presented in this work, resulted in the generation of 240 grams of CO2 emissions. The modest emission levels can be attributed to the limited size of the dataset, comprising approximately 3,000 instances, which consequently led to a reduced computational runtime during the training phase. Furthermore, the utilization of the comparatively resource-efficient ClimateBert models played a pivotal role in ensuring minimal carbon emissions throughout the experimental procedures.


\begin{table}[h!]
    \centering
    \begin{tabular}{
        lc
    }
    \toprule 
    \textbf{Task} & \textbf{CO2 emissions (kgs)} \\
    \midrule 
    Hyperparameter tuning & 0.18 \\
    Training & 0.06 \\
    \midrule 
    \textbf{Total} & \textbf{0.24} \\
    \bottomrule 
    \end{tabular}
    \caption{Estimated CO2 emissions produced during the experimental phase of the study, broken down by task.}
    \label{table:carbon-emissions}
\end{table}

\section{Bibliographical References}\label{sec:reference}
\bibliographystyle{lrec-coling2024-natbib}
\bibliography{lrec-coling2024-example}
\section{Language Resource References}
\label{lr:ref}
\bibliographystylelanguageresource{lrec-coling2024-natbib}
\bibliographylanguageresource{languageresource}

\end{document}